\documentclass[11pt,a4paper]{article}
\usepackage[hyperref]{acl2019}
\usepackage{times}
\usepackage{latexsym}
\usepackage{color}
\usepackage{hhline}

\usepackage{url}
\usepackage{breakurl}

\aclfinalcopy 
\def\aclpaperid{1446} 

\title{Translating Translationese: A Two-Step Approach to Unsupervised Machine Translation}

\author{Nima Pourdamghani$^\clubsuit$ \enspace Nada Aldarrab$^\spadesuit$ \enspace Marjan Ghazvininejad$^\diamondsuit$ \\ \textbf{Kevin Knight}$^{\heartsuit}$ \enspace \textbf{Jonathan May}$^\spadesuit$\\
  $^\clubsuit$ Amazon \enspace 
  $^\spadesuit$ USC Information Sciences Institute  \\
  $^\diamondsuit$ Facebook AI Research  \enspace
  $^\heartsuit$ DiDi Labs\\
  {\tt nimpourd@amazon.com aldarrab@isi.edu ghazvini@fb.com}\\ 
  {\tt kevinknight@didiglobal.com jonmay@isi.edu} 
  }

\date{}

\begin{document}
\maketitle
\begin{abstract}
  Given a rough, word-by-word gloss of a source language sentence, target language natives can uncover the latent, fully-fluent rendering of the translation. In this work we explore this intuition by breaking translation into a two step process: generating a rough gloss by means of a dictionary and then `translating' the resulting pseudo-translation, or `Translationese' into a fully fluent translation. We build our Translationese decoder once from a mish-mash of parallel data that has the target language in common and then can build dictionaries on demand using unsupervised techniques, resulting in rapidly generated unsupervised neural MT systems for many source languages. We apply this process to 14 test languages, obtaining better or comparable  translation results on high-resource languages than previously published unsupervised MT studies, and obtaining good quality results for low-resource languages that have never been used in an unsupervised MT scenario. 

\end{abstract}

\section{Introduction}

Quality of machine translation, especially neural MT, highly depends on the amount of available parallel data. For a handful of languages, where parallel data is abundant, MT quality has reached quite good performance \cite{wu2016google,hassan2018achieving}. However, the quality of translation rapidly deteriorates as the amount of parallel data decreases \cite{koehn2017six}. Unfortunately, many languages have close to zero parallel texts. Translating texts from these languages requires new techniques. 

\newcite{hermjakob-EtAl:2018:Demos} presented a hybrid human/machine translation tool that uses lexical translation tables to gloss a translation and relies on human language and world models to propagate glosses into fluent translations. Inspired by that work, this work investigates the following question: Can we replace the human in the loop with more technology? We provide the following two-step solution to unsupervised neural machine translation:

\begin{enumerate}
    \item \label{step:bidir} Use a bilingual dictionary to gloss the input into a pseudo-translation or `Translationese'.
    \item Translate the Translationese into target language, using a model built in advance from various parallel data, with the source side converted into Translationese using Step~\ref{step:bidir}. 
\end{enumerate}

The notion of separating adequacy from fluency components into a pipeline of operations dates back to the early days of MT and NLP research, where the inadequacy of word-by-word MT was first observed \cite{yngve55,oswald1952word}. A subfield of MT research that seeks to improve fluency given disfluent but adequate first-pass translation is \textit{automatic post-editing} (APE) pioneered by \newcite{knightchander94}. Much of the current APE work targets correction of black-box MT systems, which are presumed to be supervised.

Early approaches to unsupervised machine translation include decipherment methods \cite{Nuhn2013,Ravi2011,pourdamghani2017deciphering}, which suffer from a huge hypothesis space. Recent approaches to zero-shot machine translation include pivot-based methods \cite{chen2017teacher,zheng2017maximum,cheng2016neural} and multi-lingual NMT methods \cite{firat2016multi,firat2016zero, johnson2016google, ha2016toward, ha2017effective}. These systems are zero-shot for a specific source/target language pair, but need parallel data from source to a pivot or multiple other languages.

More recently, totally unsupervised NMT methods are introduced that use only monolingual data for training a machine translation system. \newcite{lample2017unsupervised,lample2018phrase}, \newcite{artetxe2017unsupervised}, and \newcite{yang2018unsupervised} use iterative back-translation to train MT models in both directions simultaneously. Their training takes place on massive monolingual data and requires a long time to train as well as careful tuning of hyperparameters.  

The closest unsupervised NMT work to ours is by \newcite{kim2018improving}. Similar to us, they break translation into glossing and correction steps. However, their correction step is trained on artificially generated noisy data aimed at simulating glossed source texts. Although this correction method helps, simulating noise caused by natural language phenomena is a hard task and needs to be tuned for every language. 

Previous zero-shot NMT work compensates for a lack of source/target parallel data by either using source/pivot parallel data, extremely large monolingual data, or artificially generated data. These requirements and techniques limit the methods' applicability to real-world low-resource languages. Instead, in this paper we propose using parallel data from high-resource languages to learn `how to translate' and apply the trained system to low resource settings. We use off-the-shelf technologies to build word embeddings from monolingual data~\cite{bojanowski2017enriching} and learn a source-to-target bilingual dictionary using source and target embeddings~\cite{lample2018word}. Given a target language, we train source-to-target dictionaries for a diverse set of high-resource source languages, and use them to convert the source side of the parallel data to Translationese. We combine this parallel data and train a Translationese-to-target translator on it. Later, we can build source-to-target dictionaries on-demand, generate Translationese from source texts, and use the pre-trained system to rapidly produce machine translation for many languages without requiring a single line of source-target parallel data.

We introduce the following contributions in this paper:
\begin{itemize}
    \item Following \newcite{hermjakob-EtAl:2018:Demos}, we propose a two step pipeline for building a rapid neural MT system for many languages. The pipeline does not require parallel data or parameter fine-tuning when adapting to new source languages.
    
    
    \item The pipeline only requires a comprehensive source to target dictionary. We show that this dictionary can be easily obtained using off-the shelf tools within a few hours. 
    
    
    \item We use this system to translate test texts from 14 languages into English.
    We obtain better or comparable quality translation results on high-resource languages than previously published unsupervised MT studies, and obtain good quality results for low-resource languages that have never been used in an unsupervised MT scenario.
    To our knowledge, this is the first unsupervised NMT work that shows good translation results on such a large number of languages.
    
\end{itemize}

\section{Method}
We introduce a two-step pipeline for unsupervised machine translation. In the first step a source text is glossed into a pseudo-translation or Translationese, while in the second step a pre-trained model translates the Translationese into target. We introduce a fully unsupervised method for converting the source into Translationese, and we show how to train a Translationese to target system in advance and apply it to new source languages. 

\subsection{Building a Dictionary}
The first step of our proposed pipeline includes a word-by-word translation of the source texts. This requires a source/target dictionary. Manually constructed dictionaries exist for many language pairs, however cleaning these dictionaries to get a word to word lexicon is not trivial, and these dictionaries often cover a small portion of the source vocabulary, focusing on stems and specifically excluding inflected variants. In order to have a comprehensive, word to word, inflected bi-lingual dictionary we look for automatically built ones. 

Automatic lexical induction is an active field of research \cite{fung1995compiling,koehn2002learning,haghighi2008learning,lample2018word}. A popular method for automatic extraction of bilingual dictionaries is through building cross-lingual word embeddings. Finding a shared word representation space between two languages enables us to calculate the distance between word embeddings of source and target, which helps us to find translation candidates for each word. 

We follow this approach for building the bilingual dictionaries. For a given source and target language, we start by separately training source and target word embeddings $S$ and $T$, and use the method introduced by \newcite{lample2018word} to find a linear mapping $W$ that maps the source embedding space to the target: $SW=T$.

\newcite{lample2018word} propose an adversarial method for estimating $W,$ where a discriminator is trained to distinguish between elements randomly sampled from $WS$ and $T,$ and $W$ is trained to prevent the discriminator from making accurate classifications. Once the initial mapping matrix $W$ is trained, a number of refinement steps is performed to improve performance over less frequent words by changing the metric of the space. 

We use the trained matrix $W$ to map the source embeddings into the space of the target embeddings. Then we find the $k$-nearest neighbors among the target words for each source word, according to the cosine distance metric. These nearest neighbors represent our translation options for that source word.

\subsection{Source to Translationese}\label{source_badeng}
Once we have the translation options for tokens in the source vocabulary we can perform a word by word translation of the source into Translationese. However, a naive translation of each source token to its top translation option without considering the context is not the best way to go. Given different contexts, a word should be translated differently. 

We use a 5-gram target language model to look at different translation options for a source word and select one based on its context.  This language model is trained in advance on large target monolingual data. 

In order to translate a source sentence into Translationese we apply a beam search with a stack size of 100 and assign a score equal to $\alpha P_{LM} + \beta d(s,t)$ to each translation option $t$ for a source token $s$, where $P_{LM}$ is the language model score, and $d(s,t)$ is the cosine distance between source and target words. We set $\alpha=0.01$ and $\beta=0.5$


\subsection{Translationese to Target}
We train a transformer model~\cite{vaswani2017attention} on parallel data from a diverse set of high-resource languages to translate Translationese into a fluent target. For each language we convert the source side of the parallel data to Translationese as described in Section~\ref{source_badeng}. Then we combine and shuffle all the Translationese/target parallel data and train the model on the result. 
Once the model is trained, we can apply it to the Translationese coming from any source language. 

We use the tensor2tensor implementation\footnote{\url{https://github.com/tensorflow/tensor2tensor}} of the transformer model with the \texttt{transformer\_base} set of hyperparameters (6 layers, hidden layer size of 512) as our translation model.

\section{Data and Parameters}

For all our training and test languages, we use the pre-trained word embeddings\footnote{\burl{https://github.com/facebookresearch/fastText/blob/master/pretrained-vectors.md}} trained on Wikipedia data using fastText~\cite{bojanowski2017enriching}. These embeddings are used to train bilingual dictionaries. 

We select English as the target language. In order to avoid biasing the trained system toward a language or a specific type of parallel data, we use diverse parallel data on a diverse set of languages to train the Translationese to English system. We use Arabic, Czech, Dutch, Finnish, French, German, Italian, Russian, and Spanish as the set of out training languages. 

We use roughly 2 million sentence pairs per language and limit the length of the sentences to 100 tokens. For Dutch, Finnish, and Italian we use Europarl~\cite{europarl} for parallel data. For Arabic we use MultiUN~\cite{multiun}. For French we use CommonCrawl. For German we use a mix of  CommonCrawl (1.7M), and  NewsCommentary (300K). The numbers in parentheses show the number of sentences for each dataset. For Spanish we use CommonCrawl (1.8M), and Europarl (200K). For Russian we use Yandex (1M), CommonCrawl (800K), and NewsCommentary (200K), and finally for Czech we use a mix of ParaCrawl (1M), Europarl (640K), NewsCommentary (200K), and CommonCrawl (160K).

We train one model on these nine languages and apply it to test languages not in this set. Also, to test on each of the training languages, we train a model where the parallel data for that language is excluded from the training data. In each experiment we use 3000 blind sentences randomly selected out of the combined parallel data as the development set. 

We use the default parameters in \newcite{lample2018word} to find the cross-lingual embedding vectors. In order to create the dictionary we limit the size of the source and target (English) vocabulary to 100K tokens. For each source token we find 20 nearest neighbors in the target language. We use a 5-gram language model trained on 4 billion tokens of Gigaword to select between the translation options for each token.  We use Moses scripts for tokenizing and lowercasing the data. We do not apply BPE~\cite{sennrich2015neural} on the data. In order to be comparable to \newcite{kim2018improving} we split German compound words only for the newstest2016 test data. We use the CharSplit\footnote{\url{https://github.com/dtuggener/CharSplit}} python package for this purpose. 
We use tensor2tensor's \texttt{transformer\_base} hyperparameters to train the transformer model on a single gpu for each language. 

\section{Experiments}

We report translation results on newstest2013 for Spanish, newstest2014 for French, and newstest2016 for Czech, German, Finnish, Romanian, and Russian. We also report results on the first 3000 sentences of GlobalVoices2015\footnote{\url{http://opus.nlpl.eu/GlobalVoices.php}} for Dutch, Bulgarian, Danish, Indonesian, Polish, Portuguese, and Catalan. In each experiment we report the quality of the intermediate Translationese as well as the scores for our full model.

\begin{table}[th!]
  \centering
    \tabcolsep=0.11cm
  \begin{tabular}{|p{2.3cm}|l|l|l|l|}
  \hline
     & fr-en & de-en & ru-en & ro-en \\ \hline
   \newcite{lample2017unsupervised} & 14.3 & 13.3 & - & -  \\ \hline
   \newcite{artetxe2017unsupervised} & 15.6 & 10.2 & - & - \\ \hline
    \newcite{yang2018unsupervised} & 15.6 & 14.6 & - & - \\ \hline
    \newcite{lample2018phrase} (transformer) & 24.2 & 21.0 & 9.1 & 19.4  \\ \hhline{|=|=|=|=|=|}
    \newcite{kim2018improving} & 16.5 & 17.2 & - & - \\ \hhline{|=|=|=|=|=|}
    Translationese  & 11.6 & 13.8 & 5.7 & 8.1 \\ \hline
    Full Model  & 21.0 & 18.7 & 12.0 & 16.3 \\ \hline
  \end{tabular}	
  \caption{Comparing translation results on newstest2014 for French, and newstest2016 for Russian, German, and Romanian with previous unsupervised NMT methods. \newcite{kim2018improving} is the method closest to our work. We report the quality of Translationese as well as the scores for our full model.}
  \label{T1}
\vspace{-0.3cm}
\end{table}

We compare our results against all the existing fully unsupervised neural machine translation methods in Table~\ref{T1} and show better results on common test languages compared to all of them except \newcite{lample2018phrase} where, compared to their transformer model,\footnote{They present better results when combining their transformer model with an unsupervised phrase-based translation model.} we improve results for Russian, but not for other languages.

The first four methods that we compare against are based on back-translation. These methods require huge monolingual data and large training time to train a model per test language. The fifth method, which is most similar to our approach \cite{kim2018improving}, can be trained quickly, but still is fine tuned for each test language and performs worse than our method. Unlike the previous works, our model can be trained once and applied to any test language on demand. Besides this, these methods use language-specific tricks and development data for training their models while our system is trained totally independent of the test language. 

We also show acceptable BLEU scores for ten other languages for which no previous unsupervised NMT scores exist, underscoring our ability to produce new systems rapidly (Table~\ref{T2}).

\begin{table}[th!]
  \centering
    \tabcolsep=0.11cm
  \begin{tabular}{|l|l|l|l|l|l|}
  \hline
       & cs-en & es-en & fi-en & nl-en & bg-en \\ \hline
      Translationese & 7.4 & 12.7 & 3.8 & 16.9 & 10.0 \\ \hline
      Full Model & 13.7 & 22.2 & 7.2 & 22.0 & 16.8\\ \hline
      &  da-en & id-en & pl-en & pt-en & ca-en\\ \hline
      Translationese & 13.6 & 7.4 & 8.3 & 15.2 & 10.1 \\ \hline
      Full Model & 18.5 & 13.7 &  14.8 & 23.1 & 19.8\\ \hline
  \end{tabular}	
  \caption{Translation results on ten new languages: Czech, Spanish, Finnish, Dutch, Bulgarian, Danish, Indonesian, Polish, Portuguese, and Catalan}
  \label{T2}
\vspace{-0.3cm}
\end{table}

\section{Conclusion}
We propose a two step pipeline for building a rapid unsupervised neural machine translation system for any language. The pipeline does not require re-training the neural translation model when adapting to new source languages, which makes its application to new languages extremely fast and easy. The pipeline only requires a comprehensive source-to-target dictionary. We show how to easily obtain such a dictionary using off-the shelf tools. We use this system to translate test texts from 14 languages into English.
We obtain better or comparable quality translation results on high-resource languages than previously published unsupervised MT studies, and obtain good quality results for ten other languages that have never been used in an unsupervised MT scenario.

\section*{Acknowledgements}

The research is based upon the work that took place in Information Sciences Institute (ISI) which was supported by the Office of the Director of National Intelligence (ODNI), Intelligence Advanced Research Projects Activity (IARPA), via  AFRL Contract FA8650-17-C-9116 and by the Defense Advanced Research Projects Agency (DARPA) via contract HR0011-15-C-0115.  The views and conclusions contained herein are those of the authors and should not be interpreted as necessarily representing the official policies or endorsements, either expressed or implied, of the ODNI, IARPA, DARPA, or the U.S. Government. The U.S. Government is authorized to reproduce and distribute reprints for Governmental purposes notwithstanding any copyright annotation thereon.

\bibliography{naaclhlt2019}
\bibliographystyle{acl_natbib}

\end{document}